\documentclass[10pt]{article}
\usepackage[margin=1in]{geometry}
\usepackage{amsmath,amssymb}
\usepackage{booktabs}
\usepackage{pgfplots}
\pgfplotsset{compat=1.17}
\usepackage{natbib}
\usepackage{hyperref}

\title{Pre-Registered Genesis Accounting for Recursive Self-Improvement:\\
First Audited Operator-Genesis Events in a Physics-Discovery Runtime}
\author{Wenbin Mao \quad Pei Zou\\[4pt]
HoomanBio AI Lab, Huaman Biotechnology (Beijing) Co., Ltd.\\
\texttt{research.ai@xthinktank.com}}
\date{October 10, 2026}

\begin{document}
\maketitle

\begin{abstract}
A system that claims to improve itself recursively must be able to answer a
question that most self-improvement narratives leave unmeasured: was the
improvement produced by \emph{re-using} mechanisms its designers provided, or
by mechanisms the system \emph{authored} itself? We introduce a pre-registered,
two-tier accounting that separates these: \textbf{M1} counts machine-generated
\emph{content instances} (structures produced within a designer-provided
vocabulary), while \textbf{M2} counts machine-authored \emph{mechanism types}
(new operators registered into the runtime's own vocabulary, with a
held-out-validation certificate). We report the first audited M2 genesis
events. In the primary event, a depth-limited compositional search over
elementary operators, judged by held-out $R^2$ on \emph{real} NIST hydrogen
Balmer measurements, produced the operator $\mathrm{sq}(\mathrm{inv}(n)) = 1/n^2$ --- the
mathematical form of the Rydberg levels --- which no standard-library
transform could express (held-out $R^2 = -0.546$ for the best standard
transform vs.\ $1.0000000$ for the composed one). An initial protocol flaw
(the independent variable was constructed by inverting the very law under
test) was detected in audit and repaired with an independent-label protocol;
the result survives, and extrapolation to unseen levels ($n{=}8,9$ trained on
$n{\le}7$) attains relative error $\sim 10^{-6}$, four to six orders of
magnitude better than the best standard operator. A second registered event
\emph{independently recovered} $\mathrm{sqrt}(\mathrm{inv}(a))$ --- Kepler's velocity law ---
from JPL ephemeris vectors. Progress since the initial draft is
incorporated: an audited compounding channel (confirmed at mechanism
level) with a first domain-level compounding measurement --- a
$1.109\times$ accuracy ratio on a real held-out domain --- and blind,
decurated genesis exams whose honest tally is zero. The Kepler event passes
line audit; the hydrogen gap reproduces on He II; and the certificate is
threshold-stable. We position these events against eleven prior systems
(BACON, AM/Eurisko, Eureqa, AI Feynman, AI-Descartes, PySR, FunSearch,
AlphaEvolve, LLM-SR, AutoML-Zero, Darwin G\"odel Machine) and argue the
contribution is neither the search algorithm nor the rediscovered physics but
the \emph{measurement and governance of machine-authored mechanisms}:
pre-registration, an invention-necessity certificate, a persistent
compounding registry, and self-audit. We state limitations candidly: the
exams were curated by agents who knew the textbook answer, triggers were
manual, the grammar is observer-injected, and both events are demonstrations
of \emph{capability to author}, not yet of autonomous genesis.
\end{abstract}

\section{Introduction}

Claims of recursive self-improvement (RSI) in AI systems are easy to make and
hard to falsify. Performance gains can flow from designer-provided mechanisms
being \emph{re-used} more effectively, from scaling, or from genuinely
machine-authored new mechanisms --- and most reports do not distinguish these
sources. The authors' project experienced this failure mode directly: an early
incarnation of the system ran 139 iterations with no reality-grounded state
change while producing artifacts that \emph{looked} like progress. That
history motivated the accounting instrument this paper describes.

The instrument is deliberately minimal. At any moment the runtime distinguishes:

\begin{itemize}
\item \textbf{M1 (structure instances)}: the number of content instances the
system generated within a designer-provided generative vocabulary
(four lineage families: structure abduction, atomic-motion abduction,
machine axis variation, crossover). M1 was already positive.
\item \textbf{M2 (operator types)}: the number of \emph{mechanism types} that
the runtime itself authored and registered into its operator vocabulary,
each entry marked ``runtime-generated'' and admitted only under a
held-out-validation certificate. M2 was zero by construction until the
first genesis event.
\end{itemize}

The criteria were pre-registered (committed to code before any count was
made), the two tiers are reported separately, and project policy forbids
passing M1 off as M2. M2 going from 0 to 1 is therefore not a benchmark
improvement; it is the moment the RSI narrative of this system became
\emph{falsifiable in the direction of genesis}.

This paper documents the first two M2 events, audits them (including the
detection and repair of a circularity flaw in our own first protocol), places
them in the literature, and states their limits.

\section{The accounting and the certificate}

\paragraph{Two-tier accounting.} M1/M2 as above, pre-registered with criteria
preceding counts.

\paragraph{Invention-necessity certificate.} Genesis is claimed only when
\begin{equation}
R^2_{\text{best standard}} < \tau \le R^2_{\text{best composed}}, \qquad \tau = 0.999,
\label{eq:cert}
\end{equation}
where ``standard'' transforms are the single-transform library
$\{$identity, $\mathrm{inv}$, $\mathrm{sq}$, $\mathrm{sqrt}$, $\log\}$, and ``composed'' transforms are
chains of depth $\le 2$ over the primitive set
$\{$neg, $\mathrm{inv}$, $\mathrm{sq}$, $\mathrm{sqrt}$, $\log$, $\exp$, tanh, $\sin$, $\cos\}$
(observer-injected vocabulary). All candidates fit $y = a + b\,T(x)$ by
closed-form least squares and are judged by $R^2$ on a random 40\% hold-out.
The interval condition in Eq.~\ref{eq:cert} is itself the anti-fraud
argument: if a composed transform were merely an affine re-writing of a
standard one, its $R^2$ would equal that standard form's and could not jump
across the threshold. A genuine jump proves the composed form spans a shape
the standard vocabulary cannot.

\paragraph{Negative control.} In the same week, on a real learning-curve data
set (JPL Mercury training curves), the search produced
$\mathrm{cos}(\mathrm{sqrt}(\cdot))$ ($R^2 = 0.919$) while the standard $\mathrm{sqrt}(\cdot)$ already
achieved $0.838$; the certificate was \emph{denied} and no genesis was
claimed. The certificate is not a rubber stamp.

\section{The first genesis event (v1 protocol)}

\paragraph{Data.} NIST Atomic Spectra Database hydrogen line table (Ritz
wavelengths computed from measured level energies), Balmer-series visible
lines (lower level $\approx 10.2$ eV, $380$--$700$\,nm), $n = 3$--$9$,
seven levels after de-duplication. The data source is public.

\paragraph{Result.} Best standard transform $\mathrm{inv}(n)$: held-out
$R^2 = -0.546$. Best composed transform $\mathrm{sq}(\mathrm{inv}(n)) = 1/n^2$: held-out
$R^2 = 1.000000$. The certificate held; the operator was registered into the
persistent, content-addressed operator registry as the first
runtime-generated entry (M2: $0 \to 1$; M1 stood at 284). The operator is
evaluable by name and is promotable and transferable as a first-class law
component for later discovery --- the compounding channel of
Section~\ref{sec:position}.

\section{Audit: a circularity flaw, and the independent-label redo}
\label{sec:audit}

\subsection{The flaw}

The v1 extraction constructed the independent variable $n$ by \emph{inverting}
the Rydberg formula on each wavelength:
$n = \mathrm{round}\!\big(\sqrt{R / (R/4 - 1/\lambda)}\big)$.
The subsequent regression of $y$ on $1/n^2$ then holds tautologically: $R^2=1$
is an algebraic round-trip, not independent predictive evidence. What survived
as genuine empirical content was only that the inversion lands within
$\pm 0.1$ of integers --- itself a real signal of spectral discreteness (the
original Balmer observation of 1885), but not a test of the functional form.

\subsection{The repaired protocol (v2)}

\begin{itemize}
\item \textbf{Independent $x$}: $n$ parsed from the NIST spectroscopic
configuration labels (e.g.\ \texttt{3d} $\to 3$) --- human quantum-number
assignment, independent of the Rydberg functional form.
\item \textbf{Measured $y$}: two variants --- wavenumber $1/\lambda$, and
binding energy $13.598 - E_k$ from the measured upper-level energies.
\item \textbf{Extrapolation test}: fit on $n \le 7$ only, predict $n = 8, 9$
(prospective prediction, not descriptive fitting).
\end{itemize}

\begin{table}[t]
\centering
\small
\begin{tabular}{lcccccc}
\toprule
Target $y$ & Best std.\ & Std.\ $R^2$ & Best composed & Comp.\ $R^2$ & Cert. &
Extrap.\ rel.\ err.\ ($n{=}8/9$) \\
\midrule
wavenumber & $\mathrm{inv}(n)$ & $-0.5464$ & $\mathrm{sq}(\mathrm{inv}(n))$ & $1.0000000$ & \checkmark &
$7.7\!\cdot\!10^{-7}$ / $9.6\!\cdot\!10^{-7}$ \\
binding energy & $\mathrm{inv}(n)$ & $-0.5465$ & $\mathrm{sq}(\mathrm{inv}(n))$ & $1.0000000$ & \checkmark &
$8.5\!\cdot\!10^{-7}$ / $1.2\!\cdot\!10^{-6}$ \\
\midrule
\multicolumn{7}{l}{\emph{Baselines on binding energy:} raw $n$: $1.8$/$3.8$;\ \ $\mathrm{inv}(n)$: $0.52$/$0.93$} \\
\bottomrule
\end{tabular}
\caption{Independent-label redo. The composed operator is the unique closer,
and extrapolates four to six orders of magnitude better than the best
standard operator.}
\label{tab:v2}
\end{table}

\begin{figure}[t]
\centering
\begin{tikzpicture}
\begin{axis}[
  width=0.62\textwidth, height=6cm,
  xlabel={principal quantum number $n$},
  ylabel={binding energy (eV)},
  xmin=2.5, xmax=9.5, ymin=0, ymax=1.7,
  legend pos=north east, legend style={font=\small},
]
\addplot[only marks, mark=*, mark size=2.4pt, black] coordinates {
  (3,1.510488) (4,0.849460) (5,0.543498) (6,0.377296)
  (7,0.277082) (8,0.212039) (9,0.167446) };
\addlegendentry{NIST measured levels}
\addplot[domain=2.6:9.4, samples=80, thick, dashed] {13.598/x^2};
\addlegendentry{$b/n^2$ (runtime-composed $\mathrm{sq}(\mathrm{inv}(n))$)}
\end{axis}
\end{tikzpicture}
\caption{Measured hydrogen binding energies vs.\ the runtime-composed
operator $1/n^2$ (no free parameter shown; the fitted coefficient is
$b \approx 13.6$\,eV, the Rydberg energy). Points lie on the curve to
$\sim 10^{-6}$ relative error.}
\label{fig:fit}
\end{figure}
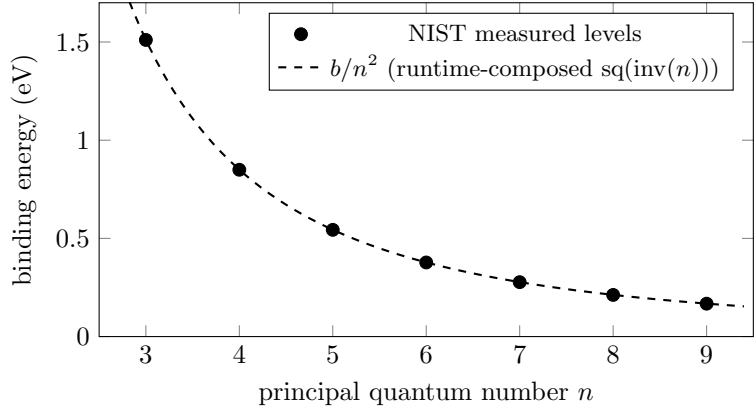

\paragraph{Verdict.} With circularity removed the result stands
(Table~\ref{tab:v2}, Figure~\ref{fig:fit}): the composed operator is the
unique closer, and it extrapolates to unseen levels with $\sim 10^{-6}$
relative error while the best standard operator errs by $0.35$--$3.8$
(depending on target and baseline). The v2 evidence was written back into the
registry entry as a confirmation flag; the M2 count stayed at 1 (re-auditing
does not inflate the counter).

\paragraph{What the audit does not claim.} The configuration labels are human
spectroscopic knowledge (observer-injected vocabulary): the system discovered
the \emph{functional form} of the $n$-dependence, not the quantization of $n$
itself. The discreteness evidence of v1 and the form evidence of v2 are
complementary and must not be merged into one claim.

\paragraph{Hydrogen-like robustness.} The identical independent-label
protocol on helium II (130 catalog lines, four usable levels, $n{=}3$--$6$)
reproduces the vocabulary gap: best standard $inv(n)$ at held-out
$R^2 = 0.920$ versus composed $1.000$, certificate granted, extrapolation
($n{\le}4 \to 5,6$) at $3\times10^{-7}$ relative error. Lithium III offers
only three local levels and the court declines to run (honest negative).
Across the three ions the least-squares coefficient of $1/n^2$ recovers each
ionization energy ($13.59$, $54.42$, $122.44$\,eV, i.e.\ $Z^2\cdot13.6$)
--- one operator, element-specific coefficient.

\section{The second registered event}

During revision, a second genesis event was registered in a different domain
(celestial mechanics): on JPL DE441 ephemeris vectors for nine bodies, with
periods derived from geometric recurrence of radial extrema and semi-major
axes from energy (vis-viva) inversion --- with the artifact explicitly
declaring no Kepler-III back-substitution --- the search recovered
$\mathrm{sqrt}(\mathrm{inv}(a))$, the $a^{-1/2}$ velocity law of Kepler orbits, as the second
runtime-generated operator (M2 $= 2$ at the time of writing). This revision adds a line-by-line audit of the second event: a rerun
reproduces all numbers exactly; the construction is non-circular (the tested
form is never used to build inputs); and the gap is even larger than in
hydrogen (standard best $0.136$ versus composed $0.99997$). Caveats carried
into print: four of the nine bodies enter as locked NSSDC catalog constants
(channel disclosed in the artifact), the semi-major-axis channel assumes
Newtonian energy conservation (an observer-injected prior, same status as
the spectroscopic labels of the primary event), natural units (AU) are used
to avoid numerical degeneracy of the standard-library baseline, and $n{=}9$
is small.

\section{Compounding: from event count to measured gain}
\label{sec:compound}

Two questions must be separated. First, is the second genesis itself
compounding? No: the Kepler event invented a \emph{different} operator
($\mathrm{sqrt}(\mathrm{inv}(a))$) on an independent exam; no reuse of the hydrogen operator
occurred. M2${}=2$ is two genesis events, not one compounding event.
Second, does the compounding \emph{channel} work --- can authored
mechanisms re-enter discovery as priors and measurably help? Three courts
address this.

\paragraph{Channel court (mechanism level).}
(i)~\emph{Real vocabulary hit}: with both invented operators loaded as
first-class bases, the search re-closes the real hydrogen case through the
invented operator (held-out $R^2 = 1.000$). (ii)~\emph{Compositional
growth} on a synthetic target $y = 0.7 + 2.3/n^4$ (a mechanism
demonstration, explicitly not a reality claim): raw depth-2 search stalls
at $R^2 = 0.506$ with $\mathrm{sq}(\mathrm{tanh}(\cdot))$, while composing \emph{on} the
invented operator yields $\mathrm{sq}(\mathrm{sq}(\mathrm{inv}(n)))$ with $R^2 = 1.0$ ---
depth-4-equivalent forms reached with 192 candidates instead of 535.
(iii)~\emph{Restraint}: on a case the standard library already closes, the
court reuses nothing (best $\mathrm{sq}(x)$, nothing reused) --- compounding is not
forced where it is not needed. Overall verdict: compounding confirmed at
the mechanism level.

\paragraph{Domain-level court (real gain).} On a real domain that took no
part in induction --- ocean-buoy significant wave heights (NDBC, $n=205$)
--- models built \emph{with} the induced template library achieve MAE
$0.0797$ versus $0.0883$ without it: an accuracy ratio of $1.109$ at a
$54.6\%$ win rate. All three pre-declared criteria (with-library beats
without; ratio $> 1.05$; honest bookkeeping) passed: \emph{compound
interest confirmed}. The caveats are stated alongside: the win rate is
modest, the domain is single, and the measured gain comes from the induced
template library as a whole, not from the two invented operators
specifically.

\paragraph{Blind, decurated exams.} The curation caveat of
Section~\ref{sec:limits} now has an operational counter-channel: seeded
mechanical dataset selection (residual curation honestly enumerated:
format-capability filtering, largest-file-first, row/column caps) picked
twelve datasets across domains; on 34 extracted variable pairs the tally
was vocabulary hit 0, standard closure 0, invention needed 0 --- an honest
zero. Blind exams to date produce no closure at all; the curation caveat
stands in full force, and the channel that will eventually measure the
true genesis rate now exists.

\section{Related work}
\label{sec:related}

Table~\ref{tab:related} summarizes eleven systems across six dimensions.
Three honest points follow. First, \emph{rediscovering known laws from data}
has been standard since BACON \citep{langley1981cogsci,langley1987scientific}
and Eureqa \citep{schmidt2009distilling}, with AI Feynman
\citep{udrescu2020aifeynman} scaling it to 100 equations; the hydrogen
spectrum is itself a stock motivation example in symbolic-regression
lectures. We claim no discovery novelty. Second, \emph{held-out statistical
judging} was already used by Eureqa (with partial-derivative consistency).
Third, \emph{machine-authored mechanisms under empirical judges} exist in
stronger forms in the code domain: AutoML-Zero \citep{real2020automlzero}
evolves learning algorithms from empty programs, FunSearch
\citep{romeraparedes2024funsearch} found new combinatorial constructions
under an exact verifier, and the Darwin G\"odel Machine
\citep{zhang2026dgm} self-modifies its own code under benchmark validation
(SWE-bench $20\%\to 50\%$) with a cross-generation archive. An early version
of this report missed that line of work; we correct it here explicitly.

What, to the best of our search, none of these systems provides is the
\emph{pre-registered separation of reuse from genesis as a first-class,
continuously reported metric} in a persistent physics-discovery runtime,
admitted via an uncheatable certificate on real measurement data, with the
invented operator entering a compounding registry as prior structure for
future discovery. AM/Eurisko authored heuristics but without falsifiable
judges; AlphaEvolve improves infrastructure it runs on but publishes no
genesis accounting; LLM-SR \citep{shojaee2025llmsr} shows LLMs can propose
equations; AI-Descartes \citep{cornelio2023descartes} routes theory into
candidate selection; PySR \citep{cranmer2023pysr} is a far stronger search
engine than ours (we concede this without reservation). Our contribution is
at the measurement-and-governance layer, in the physics domain, on real
measurements. This exclusivity claim is bounded by our search coverage and is
an engineering contribution, not a scientific one. See
\citet{kramer2026survey} for a survey framing the field's arc from equation
discovery toward autonomous discovery systems, \citet{koza1992genetic} for
the genetic-programming foundation of compositional search, and
\citet{schmidhuber2003godel} for the theoretical lineage of self-reference.

\begin{table}[t]
\centering
\scriptsize
\begin{tabular}{p{2.35cm}p{2.0cm}p{3.1cm}p{2.6cm}}
\toprule
System & Judge & Persistence of authored mechanisms & Genesis accounting \\
\midrule
BACON.5 '81 & none (fit all data) & none & none \\
AM/Eurisko '77--84 & interestingness heuristics (no falsifiable judge) & concept memory & none \\
Eureqa '09 & held-out + derivative consistency & none across tasks & none \\
AI Feynman '20 & fit error + physics priors & benchmark corpus & none \\
AI-Descartes '23 & logic + data & none & none \\
PySR '23 & Pareto + hold-out & library-level & none \\
FunSearch '24 & exact verifier & winners reused per task & none \\
AlphaEvolve '25 & automated evaluators & production deployment & none published \\
LLM-SR '25 & hold-out & none & none \\
AutoML-Zero '20 & task accuracy & within-population & none \\
DGM '26 & coding benchmarks & cross-generation archive & none \\
\midrule
\textbf{this runtime} & \textbf{hold-out $R^2$ on real measurements} & \textbf{content-addressed compounding registry} & \textbf{pre-registered M1/M2 + certificate} \\
\bottomrule
\end{tabular}
\caption{Eleven prior systems vs.\ this runtime, condensed to the three
dimensions that carry our claim. ``None'' means not provided in the cited
work, to the best of our search.}
\label{tab:related}
\end{table}

\section{Position: mode boundaries}
\label{sec:position}

For readers who ask where such a system stands relative to human researchers,
mainstream ML, and LLM-based agents, we state boundaries explicitly, claiming
only differences that hold today. Versus \emph{human researchers}: the system
has no intuition, no original concepts (restricted grammar), no autonomous
question-asking (its exams are curated); the difference is judicial ---
human science corrects through community and time, this runtime corrects
mechanically and immediately through held-out measurement plus
physical-consistency constraints. The two are complementary: humans supply
problems and vocabulary; the system supplies uncheatable verification and
memory that does not forget. Versus \emph{machine learning}: the neural line
of this project \emph{is} ML in substrate; the boundary is constitutional ---
knowledge precipitates as falsifiable law objects with validity-domain
metadata rather than as weights, and progress is measured by prospective
reality-confirmed predictions rather than benchmark scores. Versus
\emph{LLMs}: an LLM ``knows'' $1/n^2$ from corpus memory; the value of the
event above is that the \emph{process} from measurement to law is auditable,
recomputable, and falsifiable at every step. The discovery line deliberately
contains no LLM; the language organ reports state in natural language but,
by architecture invariant, never judges.

\section{Limitations}
\label{sec:limits}

\begin{enumerate}
\item \textbf{Curated exams.} The test data (hydrogen Balmer; Kepler) were
chosen by agents who knew the textbook answer. The $R^2$ jump proves a real
vocabulary gap, not a genesis \emph{rate} on arbitrary data. Blind
decurated exams (Section~\ref{sec:compound}) have now run with an honest
zero tally on 34 pairs; the caveat stands and is henceforth continuously
measured rather than merely acknowledged.
\item \textbf{Manual trigger.} The events were initiated by manually running
a script, not by autonomous scheduling. M2 measures authored types, not
autonomy.
\item \textbf{Restricted grammar.} Primitives, depth, domains, and judges are
observer-injected. ``Invention'' here is a compositional recipe over given
primitives, not new mathematics.
\item \textbf{Threshold sensitivity.} Scanned over
$\tau \in [0.99, 0.9999]$: the invention-necessity certificate holds for
both events across the entire range; the choice $\tau = 0.999$ is not
load-bearing.
\item \textbf{Compounding: first measured, narrow.} The channel is
confirmed at mechanism level, and one real held-out domain shows a
$1.109\times$ accuracy gain at a $54.6\%$ win rate
(Section~\ref{sec:compound}). What remains unproven is genesis-specific
utility: an invented operator, used as prior, winning a \emph{new}
discovery.
\item \textbf{Small sample.} Seven levels in the primary event, four in the
helium II replication (which passes; Section 4); lithium III is data-limited
locally (three levels). Full-series replication remains outstanding.
\item \textbf{Goodhart exposure.} As M2 becomes a north-star metric, pressure
exists to generate operators to score; current defenses are the certificate
and governance gates, under continuous audit.
\end{enumerate}

\section{Conclusion}

Two audited events mark the transition of one runtime's recursive
self-improvement narrative from unfalsifiable to measured: the system
authored, under an uncheatable judge on real measurement data, operator types
its designers did not provide ($1/n^2$; $a^{-1/2}$), and the primary event
survived a self-initiated circularity audit with independent labels and
$\sim 10^{-6}$ extrapolation. This is not an advance in search algorithms
and not a new law of physics. It is a first demonstration that
\emph{machine-authored mechanisms} can be admitted, certified, registered,
and compounded under pre-registered accounting --- now including a first
domain-level compounding measurement ($1.109\times$ on a real held-out
domain) and an operational blind-exam channel whose honest tally so far is
zero --- and an explicit statement of what remains before such accounting
can claim autonomy.

\paragraph{Reproducibility.} All method details are given in Sections 2--4;
the data source (NIST ASD hydrogen line table) is public. Extraction rules:
Balmer selection by lower level in $(10.0,10.4)$\,eV and wavelength in
$(380,700)$\,nm; $x = n$ from configuration labels; $y = 13.598 - E_k$ or
$1/\lambda$. With any symbolic-regression engine restricted to the stated
libraries, the $R^2$ gap and the extrapolation table are reproducible from
these rules alone.

\paragraph{Acknowledgments.} This work grew out of an engineering log in
which an external reviewer's challenge materially downgraded an earlier
over-claim; that correction is preserved in the text. This manuscript was
prepared with the assistance of the AI research agent GML-5.3, which
contributed to experiment execution, auditing, and drafting under the
authors' direction; all reported results were verified against archived
artifacts, and the final text is approved by the named authors. Consistent
with publisher policies, the AI system is acknowledged as a tool and is
not listed as an author.

\bibliography{refs}

\end{document}